%% file: Main-Paper-WSC-Conference.tex
\documentclass{wscpaperproc}
\usepackage{latexsym}
\usepackage{graphicx}
\usepackage{mathptmx}
\usepackage[T1]{fontenc}

\usepackage{float}
\usepackage{amsmath}
\usepackage{amsfonts}
\usepackage{amssymb}
\usepackage{amsbsy}
\usepackage{amsthm}
\usepackage{xcolor}
\usepackage{subcaption}
\usepackage{enumitem}
\usepackage{amsthm}
\usepackage{appendix}
\usepackage[ruled,vlined]{algorithm2e}
\usepackage{multirow}

\usepackage[pdftex,colorlinks=true,urlcolor=blue,citecolor=black,anchorcolor=black,linkcolor=black]{hyperref}

\newtheoremstyle{wsc}% hnamei
{3pt}% hSpace abovei
{3pt}% hSpace belowi
{}% hBody fonti
{}% hIndent amounti1
{\bf}% hTheorem head fontbf
{}% hPunctuation after theorem headi
{.5em}% hSpace after theorem headi2
{}% hTheorem head spec (can be left empty, meaning `normal')i

\theoremstyle{wsc}

\begin{document}

%
% AUTHOR: AUTHOR NAMES GO HERE
% FORMAT AUTHORS NAMES Like: Author1, Author2 and Author3 (last names)
%
%		You need to change the author listing below!
%               Please list ALL authors using last name only, separate by a comma except
%               for the last author, separate with "and"
%

% setting up general page style
\pagestyle{fancyplain}

% setting up page style of first page
\thispagestyle{plain}
\firstPageHead{}

% setting up running header (authors) of subsequent pages
\chead{\fancyplain{}{\itshape Wang, Sheikh, Le, and Bidkhori}}

% setting up seperation parameters
%\headsep=72pt
\rhead{}
\cfoot{}
\renewcommand{\headrulewidth}{0pt} % (renewcommand needed in fancyhdr to remove top decorative line)
%\headrulewidth=0pt  % ("setlength" needed in fancyheading to remove top decorative line)

\input{wscbib.tex}           % Set up BiBTeX macros

% needed to make the tex document look more like the word counterpart :-(
\setlength{\baselineskip}{12.7pt}

% AUTHOR: Enter the title, all letters in upper case
\title{ Task-Aware Multi-Expert Architectures for Lifelong Deep Learning}

% AUTHOR: Enter the authors of the article, see end of the example document for further examples
\author{\begin{center}Jianyu Wang\textsuperscript{1}, Jacob Nean-Hua Sheikh\textsuperscript{2}, Cat P. Le\textsuperscript{3}, and Hoda Bidkhori\textsuperscript{1}\\
[11pt]
\textsuperscript{1}Dept.~of Computational and Data Sciences, George Mason University, Fairfax, VA, USA\\
\textsuperscript{2}Dept.~of Computer Science, George Mason University, Fairfax, VA, USA\\
\textsuperscript{3}Dept.~of Electrical and Computer Engineering, Duke University, Durham, NC, USA
\end{center}
}

\maketitle

\vspace{-10pt}

\section*{ABSTRACT}

Lifelong deep learning (LDL) enables neural networks to learn continuously across tasks while preserving prior knowledge. We propose Task-Aware Multi-Expert (TAME), a novel algorithm that leverages task similarity to guide expert model selection and knowledge transfer. TAME maintains a pool of pretrained neural networks, activating the most relevant expert for each new task. A shared dense layer integrates the selected expert’s features to generate predictions. To mitigate catastrophic forgetting, TAME employs a replay buffer storing representative samples and embeddings from past tasks, enabling continual reference. An attention mechanism further prioritizes the most relevant stored knowledge for each task. Overall, the TAME algorithm supports flexible and adaptive learning across diverse scenarios. Experiments on CIFAR-100–derived classification tasks show that TAME improves classification accuracy while maintaining performance on earlier tasks, demonstrating its effectiveness in balancing adaptation and retention in evolving task sequences.

\section{INTRODUCTION}

Lifelong deep learning is a rapidly evolving area focused on empowering deep neural networks to continually learn new tasks while retaining previously acquired knowledge and adapting to novel situations. Unlike traditional ensemble learning, which aims to improve accuracy by combining multiple models for the same task, and multi-task learning (MTL), which learns multiple tasks jointly, lifelong deep learning emphasizes continual adaptation and knowledge retention as tasks arrive sequentially. A key challenge in this domain is catastrophic forgetting, where models lose performance on earlier tasks when trained sequentially—an issue particularly evident in online or real-time systems. For instance, a traffic sign detection system may degrade in recognizing previously learned signs when updated with new ones. While current methods often employ replay mechanisms or regularization strategies, these solutions may struggle to scale in complex environments or fully exploit shared knowledge across tasks. To overcome these limitations, we propose a novel framework that dynamically integrates relevant insights from multiple models, enhancing both knowledge retention and task adaptability. This is especially advantageous in simulation-based systems, where ongoing adaptation to changing conditions is essential.

In this paper, we present TAME (Task-Aware Multi-Expert), a lifelong learning algorithm designed to enable multiple deep learning models to incrementally and collaboratively learn over time. Central to our approach is the use of task similarity to inform expert selection and guide knowledge transfer. By identifying the relationships between the current task and previously encountered ones, TAME enables selective reuse of relevant knowledge, thereby improving learning efficiency and mitigating redundancy. For example, when a new task closely resembles a prior one, the most appropriate expert model is selected to extract transferable features, promoting effective adaptation while reducing forgetting. To further address the challenge of catastrophic forgetting, TAME incorporates a finite-capacity replay buffer that stores representative samples and their corresponding feature embeddings from earlier tasks. This allows the system to revisit and reinforce prior knowledge during training. We also introduce a scaled dot-product attention mechanism that dynamically prioritizes the most relevant past experiences in the buffer. To support long-range dependency learning, the replay buffer retains representative samples from all previous tasks using a balanced replacement strategy, ensuring early task information remains accessible. While attention highlights relevant memories, the buffer preserves them—enabling sustained generalization and temporal continuity. The key contributions of this work are as follows:

\begin{itemize}
 \item We propose a Task-Aware Multi-Expert lifelong learning algorithm designed to improve adaptability and knowledge retention across sequential tasks. The model architecture consists of multiple pretrained neural networks interconnected with a shared dense layer (\textit{SDL}) for the final output. When a new task arrives, TAME uses task similarity—measured via \textit{Fréchet Inception Distance (FID)} or \textit{Cosine Similarity}—to select the most appropriate model. This model extracts feature representations from the new task, which are then stored in a replay buffer and passed to the \(SDL\) for learning while preserving the expert models to prevent forgetting,

    \item We further enhance the algorithm to incorporate a scaled dot-product attention mechanism when our model retrieves knowledge from the replay buffer. Instead of treating all stored features equally, we compute attention scores between the current task and stored past tasks to dynamically prioritize more relevant experiences. This targeted weighting of memory improves feature reuse and further enhances performance and knowledge retention.

    \item We design a tailored experimental setup using multiple expert architectures connected to a shared dense layer, applied to binary classification tasks derived from the CIFAR-100 dataset. Our framework supports controlled evaluation across multiple task sequences, enabling consistent assessment of task similarity-based strategies in high-dimensional image domains. We have also conducted an experiment to compare TAME and a classical multi-task learning approach, the Shared-Bottom model. Our results show that our proposed algorithm, compared with baselines that do not utilize task similarity, performs better in terms of Average Forgetting and Average AUROC score. 
\end{itemize}

TAME can be adapted to digital twins and traffic simulation by enabling real-time adaptation to evolving conditions. In smart manufacturing, it allows digital twins to incrementally learn new equipment behaviors—such as emerging faults or wear patterns—by selecting relevant expert models based on sensor data similarity, supporting predictive maintenance without retraining. In traffic simulation, TAME can be adapted to changing dynamics across regions or times of day by reusing expert models trained on similar patterns. TAME is also compatible with datasets from stochastic simulations, where each scenario is treated as a task, making it well-suited for simulation-informed digital twins and other real-time environments.

\section{RELATED WORK}

\textit{Lifelong deep learning}, also referred to as \textit{continual deep learning}, enables models to acquire new knowledge over time while preserving previously learned information. Unlike traditional deep learning models trained on fixed datasets, LDL models must handle a sequence of tasks and adapt dynamically. A fundamental challenge in this setting is \textit{catastrophic forgetting}, where learning a new task disrupts previously acquired knowledge~\shortcite{parisi2021survey}. To address this issue, five primary classes of methods have been proposed: regularization-based, replay-based, architectural expansion, optimization, and representation methods \shortcite{wang2024continual}. Regularization-based methods constrain weight updates to retain previously important parameters (e.g. Elastic Weight Consolidation~\shortcite{kirkpatrick2017ewc}). Replay-based methods intelligently store examples from earlier tasks and interleave them with new data during training (e.g. Experience Replay~\shortcite{shin2017continual}). Architectural expansion approaches, such as Dynamically Expandable Networks~\shortcite{yoon2018dynamic}, allocate new capacity for new tasks. Representation methods explore how the feature vectors generated by networks might be shared across tasks. Optimization methods explore the objective function used to train the network, and how this function might best prevent catastrophic forgetting.

\textit{Task Similarity} is a quantitative measure of how similar two different tasks are by some specific metric. Task similarity plays an important role across machine learning literature, such as in few-shot learning and meta-learning ——~\shortcite{le2022taskaffinity,le2022fisher,kotter2024tasksimilarity}, multi-task learning ~\shortcite{shui2019principled}, and causal inference~\shortcite{aloui2023transfer}. Task similarity-based approaches have been shown to improve knowledge transfer between tasks in deep learning applications ~\shortcite{upadhyay2024sharing}.
Task similarity can be measured in a variety of ways, depending on the specific domain. In this paper, we use Fréchet Inception Distance (FID), and Cosine Similarity~\shortcite{heusel2017gans}.
Recent studies have explored the role of task similarity in lifelong learning, particularly focusing on similarities in weight differences between teacher and student networks. For instance, the teacher-student continual learning framework proposed in ~\shortciteN{lee2021continual} employs a two-layer nonlinear neural network, where task similarity
is measured based on the models' weights. In this setup, the distance between tasks is quantified by comparing the weight differences between the teacher and student models. Their findings reveal that catastrophic forgetting is most pronounced when tasks exhibit intermediate similarity.  Additionally, the weight regularization-based task similarity framework proposed in~\shortciteN{hiratani2024disentangling} demonstrates an improvement in knowledge retention in low-dimensional latent structures. Notably, when readout similarity (i.e., similarity in weights from hidden to output layers) is low, even high feature similarity (i.e., similarity in weights from input to hidden layers) can lead to interference during learning—causing the network to misuse shared representations across tasks. While these approaches provide valuable insight for shallow neural networks, they fall short in capturing the complexities of deeper architectures and do not generalize well in real-world applications where multiple models collaborate simultaneously at large scales.

In this work, our primary objective is to demonstrate the impact of leveraging task similarity in lifelong deep learning. Our task-aware multi-expert framework employs feature-based similarity measures along with attention-guided replay to dynamically select and reuse relevant expert models. We show that incorporating task similarity significantly enhances adaptive knowledge retention across tasks in lifelong learning settings.

\section{PROPOSED METHOD}

In this section, we introduce the Task-Aware Multi-Expert (TAME) lifelong learning algorithm and its attention-enhanced algorithm (AE-TAME). Here, the proposed architecture consists of \(n\) Convolutional Neural Networks \((CNNs)\), each interconnected via a shared fully connected component, referred to as the \textit{Shared Dense Layer (SDL)}. A detailed overview of the proposed framework is provided in Figure~\ref{fig:task_aware_algorithm}.

\begin{figure}[t]
    \centering
    \includegraphics[width=0.95\linewidth]{images/TAME_Procedure.png}
    \caption{\textbf{Overview of TAME.} \textbf{Left:} \textit{TAME algorithm procedure:} a) Each \(CNN\) is pretrained on an initial task. b) When a new task arrives, its similarity to initial tasks is computed. The most similar \(CNN\) is selected. c) The selected \(CNN\) extracts features of the new task, stores them in the replay buffer, and—if past data is available—feeds both new and stored features to the \(SDL\) for classification. \textbf{Right:} \textit{Model Architecture:} Multiple \(CNN\)s connect to a shared dense layer (\(SDL\)), which performs the final classification.}

    \label{fig:task_aware_algorithm}
\end{figure}

\subsection{Task-Aware Multi-Expert Lifelong Learning Algorithm}
We present the algorithm in the context of image classification. TAME is designed to address the challenge of lifelong learning by selectively reusing pretrained expert models based on task similarity. Its architecture consists of three core components:
\vspace{-2mm}
\begin{itemize}
    \item A pool of expert models (\(CNN_i\)), each pretrained on a distinct source task.
    \item A shared dense layer \((SDL)\) that receives extracted features from the selected expert and performs task-specific classification.
    \item A fixed-capacity replay buffer that stores representative samples and their embeddings from previously seen tasks to mitigate catastrophic forgetting.
\end{itemize} 

The \(i^{th} \) \(CNN\), \(CNN_i \; , 0<i \leq n\), is initially pretrained on its corresponding initial task \(i\). Following this pretraining phase, the model enters a lifelong learning phase, where it is sequentially exposed to new tasks.

When a new task \( T_t \) arrives, the algorithm computes its similarity to each initial task \( \hat{T}_i \) and selects the expert \(CNN_i\) associated with the most similar initial task. The selected \(CNN\) generates a task-specific feature representation \( \beta(T_t, i^*) \), which is used for downstream prediction and stored in a fixed-capacity replay buffer for future reference. Unlike traditional ensemble methods, only the shared dense layer (\(SDL\)) is updated during lifelong learning, while the expert \(CNNs\) remain fixed, allowing the system to accumulate and preserve expert knowledge efficiently.
The algorithmic details are described as follows:
\vspace{-2mm}
\begin{enumerate}
  \setlength{\itemsep}{2pt} 
  \setlength\itemindent{-1em}
    \item \textbf{Initialization Phase.} During this phase, we initialize \(n\) expert \(CNNs\) and a Shared Dense Layer (\(SDL\)) with random parameters. A replay buffer is also initialized as a dictionary indexed by \(CNN_i\), where \(i = 1, \dots, n\), with each entry storing image samples, labels, and feature representation. Each \(CNN_i\) is independently pretrained on a distinct initial task \(\hat{T}_i\), and its parameters are frozen afterward to preserve learned representations.

    \item \textbf{Lifelong Learning Phase.} 
    When a new task $T_t$ arrives at time step $t$, the algorithm executes the following steps:
    \begin{itemize}   
    \setlength{\itemsep}{2pt} 
    \setlength\itemindent{-1em}
        \item {\it Step 1: Task Similarity-Based Model Selection.} 
        TAME computes the similarity between the incoming task $T_t$ and all initial tasks $\hat{T}_1, \dots, \hat{T}_n$. This is done using metrics such as: \textit{Fréchet Inception Distance (FID)} and \textit{Cosine Similarity}. FID quantifies the distance between the feature distributions of the current task and the previously learned initial tasks, where lower values indicate higher similarity. Cosine similarity, in contrast, measures the cosine of the angle between two feature vectors, with higher values indicating greater similarity. The expert model $\text{CNN}_{i^*}$ associated with the most similar source task is selected. The chosen model is used to generate a feature embedding $\beta(T_t, i^*)$ for the new task. Full definitions and implementation details of both measures are provided in Appendix~\ref{app:FID} and Appendix~\ref{app:cosine}.
        \item {\it Step 2: Replay Learning.} We employ a replay buffer as a memory module with a fixed capacity. It removes the oldest entries once full to retain recent information. As \(T_t\) arrives at time \(t\), its feature representation \(\beta (T_t, i^*)\), raw images \(Img(T_t)\) and task label \(L_t\) are stored under the selected \(CNN_{i^*}\). If replay data is available, the buffer retrieves all previous feature representations \(\beta(T_j, i^*)\) using \(CNN_i^{*}\), and their raw images \(Img(T_j) \; , 1 \leq j \leq t-1\). It concatenates them with \(\beta(T_t, i^*)\) and \(Img(T_t)\), respectively to form combined feature vector \(\theta_i\) and \(Img\): 
\[
                {\theta_i} =
                \begin{bmatrix}
                    \beta(T_t, i^*) \\
                    \beta(T_j, i^*)\; , 1 \leq j \leq t-1
                \end{bmatrix} ,
                {Img} =
                \begin{bmatrix}
                    {Img(T_t)} \\
                    {Img(T_j)}
                \end{bmatrix} ,
\] which—along with the label of all stored task \(L_j\; ,1 \leq j \leq t-1\)  and the new task label \(L_t\)—is used to update the parameters of the \(SDL\) via backpropagation. If \(T_t\) is the first task for \(CNN_{i^*}\), only \(Img(T_t)\), label \(L_t\) and feature \(\beta(T_t, i^*)\) are used, refer to Algorithm~\ref{alg:TAME_LDL}.
    \end{itemize}

\end{enumerate}

\vspace{1mm}

\begin{algorithm}[H]
\caption{Task-aware multi-expert lifelong deep learning algorithm (TAME).}
\label{alg:TAME_LDL}

\SetKwInput{KwInput}{Input}         % Set the Input
\SetKwInput{KwOutput}{Output}       % set the Output
\SetKwInput{KwData}{Data}
\SetKwFunction{initialization}{Initialization}
\SetKwFunction{lifelong}{LifelongLearning}
\SetKwComment{Comment}{$\triangleright$ }{}
\SetCommentSty{scriptsize}

\DontPrintSemicolon

\KwIn{\textit{Initialization tasks:} $\hat T_1, \dots, \hat T_n$; \textit{Lifelong tasks:} $T_1, \dots, T_t$}
\KwOut{Classification decision $O$}

\SetKwProg{Fn}{Function}{:}{} \Fn{\initialization}{ 
Initialize a replay buffer \\
\For{$\hat T_i\; , 1 \leq i \leq n $}{
    Train $CNN_i$ with $\hat T_i$ \\
    Freeze $CNN_i$\;
}
\KwRet{$CNN$}
}

\SetKwProg{Fn}{Function}{:}{} \Fn{\lifelong}{ 
    \For{incoming task $T_t$}{
        Compute similarity $d(T_t, \hat T_i)$ for all $i$, $1 \leq i \leq n$ \Comment*[r]{Step 1: Task Similarity-Based Selection}\;
        }
    Select $CNN_{i^*}$ with highest similarity to generate feature $\beta(T_t, i^*)$ \;
    Store the raw images of $T_t$, denoted as $Img(T_t)$, new task label $L_t$, and $\beta(T_t, i^*)$ in replay buffer
    
    \eIf{replay data is available \Comment*[r]{Step 2: Replay-Based Learning}}{
        Retrieve stored features $\beta(T_j, i^*)$ from the replay buffer for $1 \leq j \leq t-1$\;
        Obtain combined feature vector $\theta_i$ and combined image vector $Img$ \; 
        Pass $Img$, new task label \(L_t\), past task labels \(L_j\;, 1 \leq j \leq t-1\), and $\theta_i$ to \(SDL\)
    }{
        Pass $Img(T_t)$, new task label \(L_t\), and $\beta(T_t, i^*)$ to \(SDL\)
    }
    \KwRet{$O$}
}
\end{algorithm}

\vspace{6mm}
To evaluate the importance of task similarity, we design a baseline model that shares the same architecture and replay mechanism as TAME algorithm. However, rather than selecting the most similar \(CNN_{i^*}\), the baseline randomly assigns each incoming task \(T_t\) to one of the \(n\) \(CNN\)s, regardless of similarity.

\subsection{Attention-Enhanced Task-Aware Multi-Expert Lifelong Learning Algorithm}

We implement an attention mechanism in the TAME algorithm, while preserving the original architecture and replay buffer, referring to the enhanced model as AE-TAME. Inspired by scaled dot-product attention~\shortcite{vaswani2017attention}, AE-TAME dynamically assigns weights to stored task features based on their relevance to the current task, enabling the model to focus on more informative representations. The attention-enhanced features are then passed to the Shared Dense Layer (\(SDL\)) for final prediction.

Attention works by comparing a \textit{Query}—the current task features—to a set of features \textit{Keys} representing past tasks, and weighting the corresponding \textit{Values} based on attention weight.

Our algorithm still consists of two phases. The \textit{Initialization Phase} and \textit{Lifelong learning phase}. The \textit{Initialization Phase} remains unchanged. In the \textit{Lifelong learning phase}, Step 1—\textit{Task Similarity-Based Model Selection}—is also unchanged. The change happen in Step 2—\textit{Replay Learning}. This step incorporates an attention mechanism when the selected \(CNN_{i^*}\) retrieves stored feature in the replay buffer to enhance knowledge retention and task adaptability. In this step, we define the standard attention components as follows:

\begin{itemize}
    \item \textit{\(Q_t\) (Query)}: The feature representation \(\beta(T_t, i^*)\) generated by  \(CNN_{i^*}\) .
    
    \item \textit{\(K_j\) (Key)}: The feature representations stored in the replay buffer for \(CNN_{i^*}\) for previous lifelong learning tasks, \(1 \leq j \leq t-1\).
    
    \item \textit{\(V_t\) (Value)}: A weighted sum of the stored task features, computed using attention weights \(\alpha_j\):
    \[
    V_t = \sum_{j=1}^{t-1} \alpha_j \cdot \beta(T_j, i^*).
    \]
    where each \(\alpha_j\) is derived from the scaled dot-product attention weight. For the details, please refer to Appendix~\ref{app:attention_weight}.
\end{itemize}

When a new task \(T_t\) arrives at time step \(t\), the selected \(CNN_{i^*}\) first generates its feature representation \(\beta(T_t, i^*)\). The attention mechanism then computes a weighted sum \(V_t\) of the stored features \(\beta(T_j, i^*)\; ,1 \leq j \leq t - 1\), retrieved from the replay buffer. These are combined to form the attention-enhanced representation \(\gamma_i\): \[
\gamma_i = \left[\beta(T_t, i^*), V_t\right].
\] Finally, \(\gamma_i\), the combined image vector \(Img\), the current task label \(L_t\), and the stored task labels \(L_j, 1 \leq j \leq t - 1\), are passed to the Shared Dense Layers (\(SDL\)) for prediction, as detailed in Algorithm~\ref{alg:AE-TAME-LDL}.

\begin{algorithm}[t]
\caption{Attention-enhanced TAME algorithm (AE-TAME).}
\label{alg:AE-TAME-LDL}

\SetKwInput{KwInput}{Input}         % Set the Input
\SetKwInput{KwOutput}{Output}       % set the Output
\SetKwInput{KwData}{Data}
\SetKwFunction{initialization}{Initialization}
\SetKwFunction{lifelong}{LifelongLearning}
\SetKwComment{Comment}{$\triangleright$ }{}
\SetCommentSty{scriptsize}

\KwIn{\textit{Initialization tasks:} $\hat T_1, \dots, \hat T_n$; \textit{Lifelong tasks:} $T_1, \dots, T_t$}
\KwOut{Classification decision $O$}

\SetKwProg{Fn}{Function}{:}{}\Fn{\initialization}{ 
\For{$\hat T_i\; , 1 \leq i \leq n $}{
    Train $CNN_i$ with $\hat T_i$ \\
    Freeze $CNN_i$
    }
\KwRet{$CNN$}
}

\SetKwProg{Fn}{Function}{:}{} \Fn{\lifelong}{ 
\For{incoming task $T_t$}{
        Compute similarity $d(T_t, \hat T_i)$ for all $i$, $1 \leq i \leq n $  \Comment*[r]{Step 1: Task Similarity-Based Selection}
        }
    Select $CNN_{i^*}$ with highest similarity score to generate feature $\beta(T_t, i^*)$ \\
    Store the raw images of $T_t$, denoted as $Img(T_t)$, new task label $L_t$, $\beta(T_t, i^*)$ in replay buffer\\

\eIf{replay data exists \Comment*[r]{Step 2: Replay-Based Learning}}{ 
    Retrieve stored features $\beta(T_j, i^*)$ from the replay buffer for $1 \leq j \leq t-1$ \\
    Compute attention weights $\alpha_j$\\
    Compute $V_t = \sum \alpha_j \cdot \beta(T_j, i^*)$\\
    Obtain combined attention-enhanced representation $\gamma_i$ \\
    Pass combined image vector $Img$, new task label \(L_t\), stored task labels \(L_j\), and $\gamma_i$ to \(SDL\)
    }{
    Pass $Img(T_t)$, new task label \(L_t\), and $\beta(T_t, i^*)$ to \(SDL\)
    }
    
\KwRet{$O$}
}
\end{algorithm}

\textbf{Baseline for Evaluation of AE-TAME.} We design a baseline model that shares the same architecture, replay mechanism, and attention mechanism as the AE-TAME algorithm, referred to as the Attention-Enhanced Baseline. The only difference is that it does not use task similarity measures. This baseline randomly assigns each incoming task \(T_t\) to one of the \(n\) \(CNN\), regardless of similarity.

\section{EXPERIMENTAL RESULTS}
In this section, we demonstrate the effectiveness of TAME and AE-TAME in comparison to baseline algorithms. As discussed in the previous section, the baseline algorithms maintain the same architectural structure. The key difference lies in the expert selection step, where the baseline methods select a \(CNN\) at random rather than leveraging task similarity for informed model selection.

\subsection{Experimental Setup and Evaluation Metrics}

In this subsection, we present the experimental setup and evaluation metrics.

\begin{itemize}
    \item {\bf Dataset and Task Formulation.} 
    In machine learning, a "task" refers to a well-defined  objective that the learning algorithm aims to optimize based on input-output mappings derived from data. For instance, in an "image classification task", the goal is to categorize images into predefined classes based on their visual features.
    In our experiment, each task is a binary classification problem. They are derived from the CIFAR-100 dataset, which contains 60{,}000 color images across 100 categories. Each initial task (for pretraining) consists of 400 images (200 per class), trained for 5 epochs. During lifelong learning phase, 10 new binary tasks are introduced incrementally, each containing 200 images. We evaluate performance across 5 distinct sequences (Sequence 1–5), each comprising 10 unseen binary tasks.

    \item {\bf Model Architecture.} For TAME, AE-TAME, and the baselines, our model architecture is the same and consists of five convolutional neural networks (\(CNN\)), each comprising three convolutional layers with ReLU activations and max-pooling, followed by two fully connected layers. Each \(CNN\) contains approximately 180K parameters. These \(CNN\)s are connected to a Shared Dense Layer (\(SDL\)), which includes five fully connected layers with ReLU activations and a sigmoid output for binary classification, totaling around 720K parameters. A replay buffer with a fixed maximum capacity of 1,000 images is employed to retain past task information for lifelong learning phase.

    \item {\bf Lifelong Learning Phase.} For each task, only the \(SDL\) is updated using Binary Cross Entropy (BCE) loss over three training epochs, with the Adam optimizer (learning rate = 0.001, batch size = 32). Task similarity is computed using either Fréchet Inception Distance (FID) or Cosine Similarity. Based on the similarity score, the most appropriate \(CNN\) is selected to process the new task. A baseline variant assigns the new tasks to a randomly selected \(CNN\) without using similarity measures.

    \item {\bf Evaluation Metrics.} We evaluate and benchmark the performance of our algorithms using two standard metrics: For each task sequence, \textit{Average AUROC (A-AUROC)} is computed by averaging the AUROC (Aread Under the Receiver Operating Characteristic Curve) scores of all tasks in the sequence, reflecting the model’s overall ability to distinguish between classes. \textit{Average Forgetting (AF)} measures knowledge degradation by averaging the difference between each task’s highest and final recorded accuracy, as detailed in Appendix~\ref{app:auroc} and Appendix~\ref{app:forgetting}.

    \item {\bf Runtime and Computational Setup.} All experiments were conducted on a GPU cluster with a single GPU featuring 14 GB of memory. Under the described experimental setup, TAME and AE-TAME each required approximately 40 minutes on average to complete \(CNN\) pretraining, model training, and evaluation per trial. Although the baseline models ran slightly faster, the difference in runtime was not significant.
\end{itemize}

\subsection{Performance and Comparison of TAME and AE-TAME and Baselines} 
Table~\ref{tab:fid_similarity} compares all algorithms utilizing the FID similarity measure. AE-TAME outperforms both Baseline and AE-Baseline in terms of Average Forgetting (AF) across all five sequences, and achieves higher Average AUROC (A-AUROC) in four out of five cases, confirming its strong performance when leveraging FID-based task similarity. TAME also outperforms Baseline and AE-Baseline in five out of five sequences for AF and in three out of five sequences for A-AUROC, demonstrating that task similarity—even without the attention mechanism—significantly improves knowledge retention and classification accuracy. Furthermore, AE-TAME surpasses TAME in three out of five sequences in terms of AF and in four out of five sequences for A-AUROC. These results collectively highlight the benefits of incorporating task similarity into the model, and further show that performance is additionally enhanced when the attention mechanism is employed.

\begin{table}[t]
\centering
\caption{Performance of all algorithms using FID similarity in terms of average forgetting and average AUROC across five task sequences.}
\label{tab:fid_similarity}
\begin{tabular}{|l||cc|cc|cc|cc|}
    \hline
    \multirow{2}{*}{\textbf{Sequence}} 
    & \multicolumn{2}{c|}{\textbf{Baseline}} 
    & \multicolumn{2}{c|}{\textbf{AE-Baseline}} 
    & \multicolumn{2}{c|}{\textbf{TAME}} 
    & \multicolumn{2}{c|}{\textbf{AE-TAME}} \\
    \cline{2-9}
    & AF & A-AUROC & AF & A-AUROC & AF & A-AUROC & AF & A-AUROC \\
    \hline
    Seq 1. & 0.0815 & 0.4869 & 0.099  & 0.4871 & 0.067  & 0.5080 & \textbf{0.0185} & \textbf{0.5580} \\
    Seq 2. & 0.0964 & 0.4982 & 0.0625 & 0.4704 & \textbf{0.0051} & 0.5114 & 0.028 & \textbf{0.5643} \\
    Seq 3. & 0.072  & \textbf{0.5350} & 0.2235 & 0.4443 & 0.06   & 0.4521 & \textbf{0.052} & 0.5167 \\
    Seq 4. & 0.075  & 0.4975 & 0.0675 & 0.4369 & 0.072 & 0.4017 & \textbf{0.019} & \textbf{0.4993} \\
    Seq 5. & 0.0777 & 0.4975 & 0.0593 & 0.4512 & \textbf{0.0536} & \textbf{0.5622} & 0.052 & 0.5022 \\
    \hline
\end{tabular}
\end{table}

Table~\ref{tab:cosine_similarity} compares all algorithms utilizing the Cosine Similarity measure. The results show that AE-TAME consistently achieves lower forgetting than both Baseline and AE-Baseline across four out of five sequences and outperforms them in A-AUROC also in four out of five cases. TAME exceeds the performance of Baseline and AE-Baseline in three out of five sequences for AF and in two out of five for A-AUROC. When comparing AE-TAME to TAME, AE-TAME demonstrates lower forgetting in three out of five sequences and achieves higher A-AUROC in five out of five. These findings indicate that while both task similarity metrics are beneficial, Cosine Similarity proves particularly effective when combined with attention, especially in enhancing AUROC. Overall, AE-TAME remains the top-performing model across both task similarity metrics and evaluation criteria.

\begin{table}[H]
\centering
\caption{Performance of all algorithms using cosine similarity in terms of average forgetting and average AUROC across five task sequences.}
\label{tab:cosine_similarity}
\begin{tabular}{|l||cc|cc|cc|cc|}
    \hline
    \multirow{2}{*}{\textbf{Sequence}} 
    & \multicolumn{2}{c|}{\textbf{Baseline}} 
    & \multicolumn{2}{c|}{\textbf{AE-Baseline}} 
    & \multicolumn{2}{c|}{\textbf{TAME}} 
    & \multicolumn{2}{c|}{\textbf{AE-TAME}} \\
    \cline{2-9}
    & AF & A-AUROC & AF & A-AUROC & AF & A-AUROC & AF & A-AUROC \\
    \hline
    Seq 1. & 0.0815 & 0.4869 & 0.099  & 0.4871 & \textbf{0.0195} & 0.5345 & 0.026 & \textbf{0.5557} \\
    Seq 2. & 0.0964 & 0.4982 & 0.0625 & 0.4704 & 0.0723 & 0.5028 & \textbf{0.0445} & \textbf{0.5807} \\
    Seq 3. & 0.072  & 0.5350 & 0.2235 & 0.4443 & \textbf{0.027} & 0.5176 & 0.076 & \textbf{0.5583} \\
    Seq 4. & 0.075  & \textbf{0.4975} & 0.0675 & 0.4369 & 0.083 & 0.4268 & \textbf{0.0095} & 0.4591 \\
    Seq 5. & 0.0777 & 0.4975 & 0.0593 & 0.4512 & 0.1012 & 0.4268 & \textbf{0.0523} & \textbf{0.5785} \\
    \hline
\end{tabular}
\end{table}

\subsubsection{Additional Illustration for Accuracy}
\begin{figure}[H]
    \centering
    % First Subfigure
    \begin{subfigure}[t]{0.48\linewidth}
        \centering
        \includegraphics[width=\linewidth]{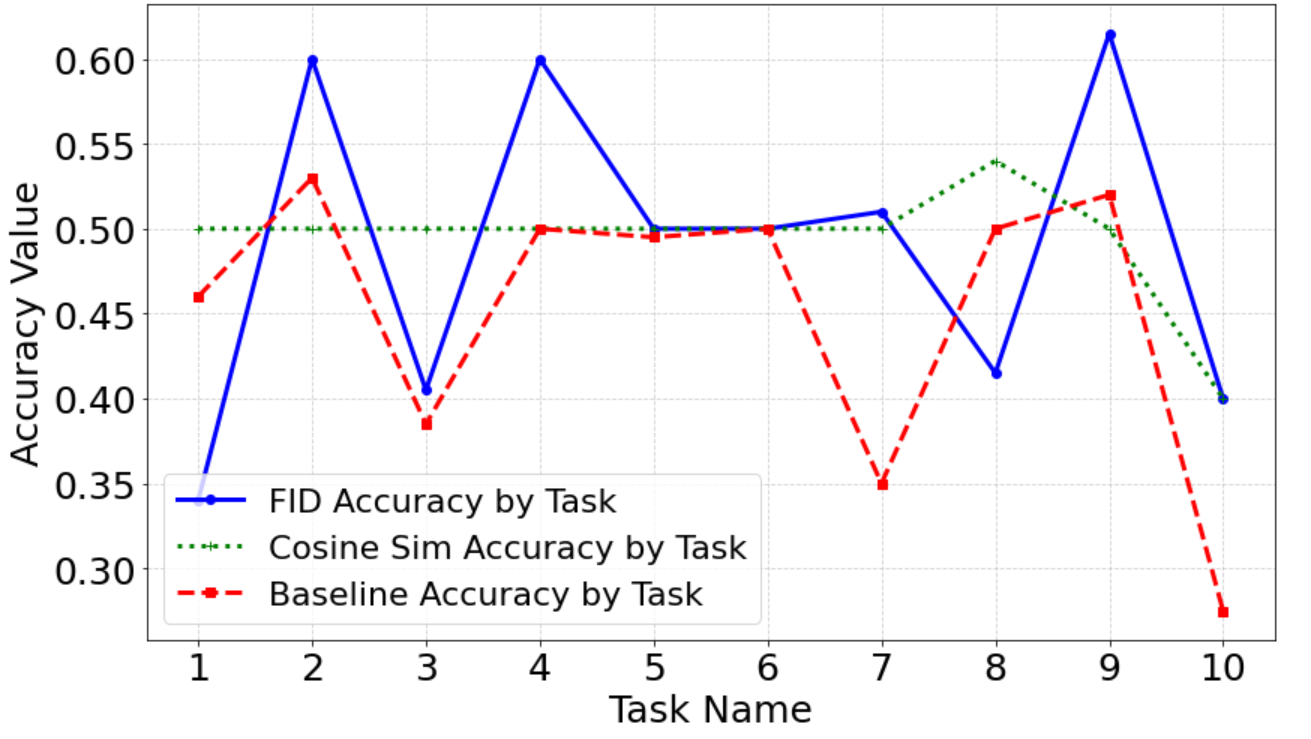}
        \caption{Accuracy across sequential tasks (TAME).}
        \label{fig:tame accuracy}
    \end{subfigure}
    \hfill
    % Second Subfigure
    \begin{subfigure}[t]{0.48\linewidth}
        \centering
        \includegraphics[width=\linewidth]{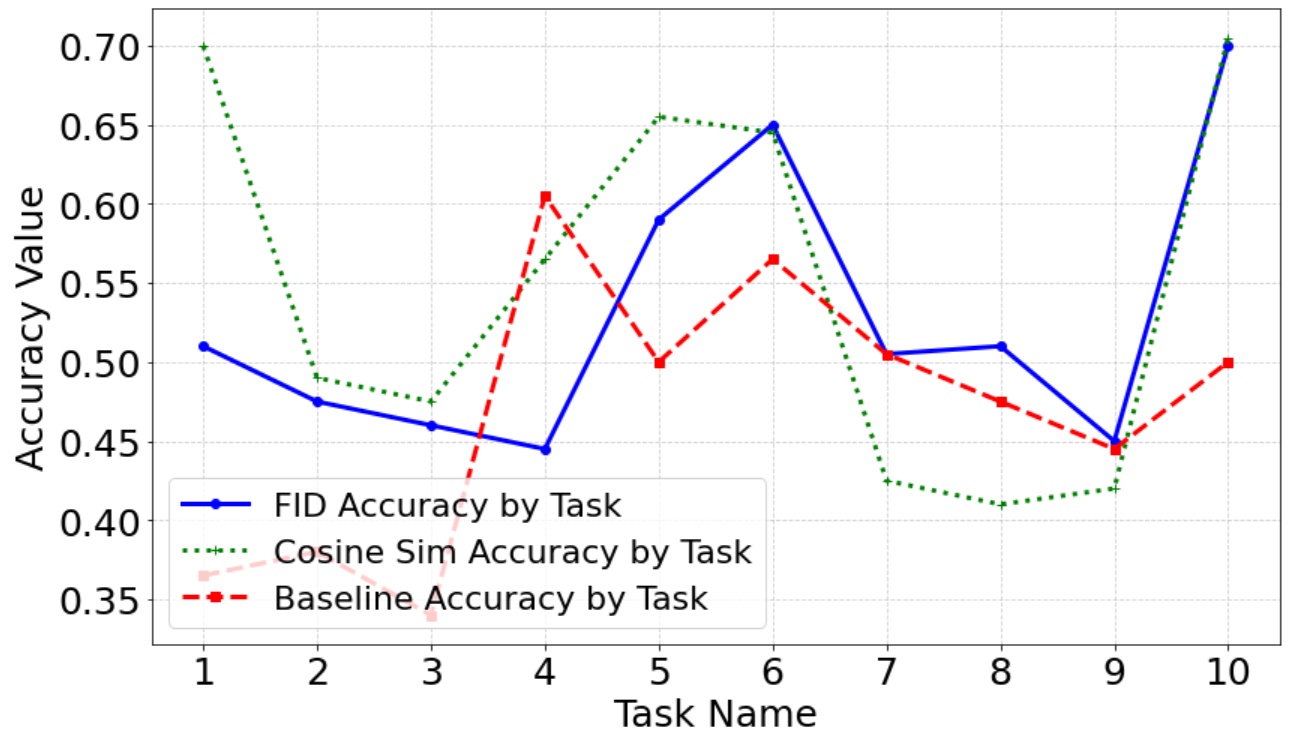}
        \caption{Accuracy across sequential tasks (AE-TAME).}
        \label{fig:ae-tame-accuracy}
    \end{subfigure}
    
    \caption{Accuracy performance of TAME and AE-TAME across tasks for one sequence from CIFAR-100.}
    \label{fig:tame ae-tame Compare}
\end{figure}

For additional illustration, Figures~\ref{fig:tame ae-tame Compare} visualize the performance of TAME and AE-TAME across one particular sequence of 10 binary classification tasks derived from CIFAR-100, showing task-specific accuracy. In Figure~\ref{fig:tame accuracy},  TAME utilizing FID and  TAME utilizing Cosine Similarity show stronger and more stable accuracy trends than the baseline. TAME utilizing FID achieves higher accuracy in later tasks, while TAME utilizing Cosine Similarity is competitive in early and middle tasks. The baseline shows lower and more volatile accuracy. In Figure~\ref{fig:ae-tame-accuracy}, AE-TAME using FID consistently outperforms the baseline across most tasks, while AE-TAME using Cosine Similarity achieves the highest accuracy, peaking around 70\%. The AE baseline remains inferior, confirming the advantage of combining attention mechanisms with task similarity for robust and consistent accuracy across tasks.

\subsection{Comparison with the Shared-Bottom model}
In this subsection, we compare the average forgetting performance of TAME and AE-TAME with a traditional multi-task learning baseline—the Shared-Bottom model \cite{zhang2022survey}. The Shared-Bottom model uses a shared \(CNN\) encoder and task-specific binary classification heads, with a structure and parameter size designed to match that of TAME. This configuration ensures a fair comparison in terms of model capacity and architectural complexity. We adopt the same experimental protocol as with TAME. The model is optimized using Adam with a learning rate of 0.001, trained for 3 epochs per task with a batch size of 32, and evaluated after each task to compute average forgetting.

As shown in Table~\ref{tab:avg_forgetting_results}, both AE-TAME and TAME completely outperforms the Shared-Bottom model across all sequences. The Shared-Bottom model, while conceptually efficient, suffers from greater forgetting due to its inability to isolate task-specific representations during continual updates.

\begin{table}[H]
\centering
\caption{Average forgetting of TAME, AE-TAME, and shared-bottom model across five task sequences.}
\label{tab:avg_forgetting_results}
\resizebox{\textwidth}{!}{%
\begin{tabular}{|l||c|c|c|c|c|}
    \hline
    \textbf{Sequence} 
    & \textbf{TAME (FID)} 
    & \textbf{TAME (Cosine)} 
    & \textbf{AE-TAME (FID)} 
    & \textbf{AE-TAME (Cosine)} 
    & \textbf{Shared-Bottom} \\
    \hline
    Seq 1. & 0.0670 & 0.0195 & \textbf{0.0185} & 0.0260 & 0.2111 \\
    Seq 2. & \textbf{0.0051} & 0.0723 & 0.0280 & 0.0445 & 0.1972 \\
    Seq 3. & 0.0600 & \textbf{0.0270} & 0.0520 & 0.0760 & 0.2083 \\
    Seq 4. & 0.0720 & 0.0830 & 0.0190 & \textbf{0.0095} & 0.1947 \\
    Seq 5. & 0.0536 & 0.1012 & \textbf{0.0520} & 0.0523 & 0.1573 \\
    \hline
\end{tabular}%
}
\end{table}

\section{CONCLUSION}
We presented a Task-Aware Multi-Expert lifelong learning framework that leverages task similarity and a multi-expert \(CNN\) architecture to facilitate incremental and collaborative learning. By dynamically selecting the most relevant expert model based on task similarity metrics, our approach enhances both prediction accuracy and knowledge retention. Additionally, the integration of an attention mechanism in AE-TAME further improves the reuse of past knowledge by focusing on the most relevant task representations. 

While this work focuses on image-based tasks and a single data modality, it can be adapted to a broad range of real-world applications involving multimodal data. TAME’s modular design makes it well suited for such extensions, enabling specialized expert models to handle different modalities. In future work, we aim to improve the scalability of the framework and incorporate task similarity to enhance other lifelong learning algorithms.

\section*{ACKNOWLEDGMENT}

This research was supported by the ONR grant N000142312629. 

\normalsize
\appendix
\section{APPENDIX}

\subsection{Fréchet Inception Distance (FID)}
\label{app:FID}
Given two tasks—Task 1 (new task) and Task 2 (initial task)—the Fréchet Inception Distance (FID) quantifies the similarity between their feature distributions by measuring the distance between two multivariate Gaussians \shortcite{heusel2017gans}. It is computed as:

\[
FID = \|\mu_1 - \mu_2\|^2 + \text{Tr}(\Sigma_1 + \Sigma_2 - 2(\Sigma_1 \Sigma_2)^{1/2}),
\]

where:
\vspace{-2mm}
\begin{itemize}
    \item \(\mu_1\), \(\mu_2\) are the mean feature vectors of Task 1 and Task 2, respectively. For a given task \(T\), the mean vector \(\mu_T\) is computed as:
    \[
    \mu_T = \frac{1}{N} \sum_{i=1}^{N} f(x_i),
    \]
    where \(f(x_i)\) is the extracted feature vector for image \(x_i\) and \(N\) is the total number of samples in the task.

    \item \(\Sigma_1\), \(\Sigma_2\) are the covariance matrices of the feature distributions of Task 1 and Task 2, respectively. Each covariance matrix \(\Sigma_T\) is calculated as:
    \[
    \Sigma_T = \frac{1}{N - 1} \sum_{i=1}^{N} (f(x_i) - \mu_T)(f(x_i) - \mu_T)^\top ,
    \]
\end{itemize}
\vspace{-2mm}

Lower FID values indicate greater similarity between the tasks, suggesting that their feature distributions are closely aligned.

\subsection{Cosine Similarity}
\label{app:cosine}
Cosine similarity quantifies the orientation between two feature vectors by measuring the cosine of the angle between them. It is computed as:

\[
\text{Cosine Similarity} = \frac{A \cdot B}{\|A\| \|B\|} ,
\]

where:
\vspace{-2mm}
\begin{itemize}
    \item \(A\): Feature vector of the new task.
    \item \(B\): Feature vector of the initial task.
\end{itemize}
\vspace{-2mm}

Higher Cosine Similarity values indicate greater alignment between the tasks, suggesting stronger task similarity.

\subsection{Attention Weight}
\label{app:attention_weight}

The attention weight \(\alpha_j\) quantifies the relevance of the \(j^{\text{th}}\) past task relative to the current task \(T_t\) and is used to prioritize informative feature representations during replay-based learning. Following the scaled dot-product attention mechanism~\shortcite{vaswani2017attention}, we normalize similarity scores between the current task and previously encountered tasks using their feature representation.

The attention weight \(\alpha_j\) is computed as:

\[
\alpha_j = \frac{\exp\left(\frac{Q_t \cdot K_j^\top}{\sqrt{d_k}}\right)}{\sum\limits_{l=1}^{t-1} \exp\left(\frac{Q_t \cdot K_l^\top}{\sqrt{d_k}}\right)} = 
\frac{\exp\left(\frac{\beta(T_t, i^*) \cdot \beta(T_j, i^*)^\top}{\sqrt{d_k}}\right)}{\sum\limits_{l=1}^{t-1} \exp\left(\frac{\beta(T_t, i^*) \cdot \beta(T_l, i^*)^\top}{\sqrt{d_k}}\right)} ,
\]

where:
\vspace{-2mm}
\begin{itemize}
    \item \( \beta(T_t, i^*) \): Feature vector (query \(Q_t\)) for the current task.
    \item \( \beta(T_j, i^*) \): Feature vector (key \(K_j\)) for the \(j^{\text{th}}\) past task.
    \item \( d_k \): Dimensionality of the feature vectors (for scaling).
\end{itemize}
\vspace{-2mm}

This attention formulation allows the model to emphasize past experiences that are most relevant to the current task, enhancing prediction quality and knowledge retention.

\subsection{AUROC}
\label{app:auroc}
AUROC score, which measures the model’s ability to distinguish between classes by calculating the area under the ROC curve. In our work, AUROC is computed for each task in a sequence, and we compute the Average AUROC by averaging these scores across all tasks

\subsection{Forgetting}
\label{app:forgetting}

Forgetting intuitively represents the degradation in a model's performance on previously learned tasks after it is trained on a new task~\cite{chaudhry2018riemannian}. Suppose we have a sequence of \(T\) tasks. For a task \(T_i\), let \(a_i^i\) denote its accuracy immediately after being learned, and \(a_t^i\) denote its accuracy after learning task \(T_t\) (for \(t > i\)). Then, the forgetting for task \(T_i\) at time \(t\) is defined as:
\[
F_t^i = \max(a^i) - a_t^i,
\]
where \(\max(a^i)\) is the highest accuracy achieved on task \(T_i\) over time.

The overall forgetting at time \(t\) is then averaged over all previously learned tasks:
\[
\text{Average Forgetting} = \frac{1}{t-1}\sum_{i=1}^{t-1} \left[\max(a^i) - a_t^i\right].
\]

\footnotesize
% AUTHOR: Include your bib file here
\bibliography{reference}

% Please don't exchange the bibliographystyle style
\bibliographystyle{wsc}

\section*{AUTHOR BIOGRAPHIES}
\noindent {\bf JIANYU WANG} is a Ph.D. student at the Department of  Computational and Data Sciences at the George Mason University. He earned his Master's degree in Information System from the Johns Hopkins University. His research interests include machine learning and artificial intelligence. His email address is \email{jwang72@gmu.edu}.

\noindent {\bf JACOB NEAN-HUA SHEIKH} 
is pursuing a dual B.S./M.S. degree in the Department of Computer Science at George Mason University, with a focus on computational neuroscience. His research interests include biologically inspired machine learning, continual learning, and applied medical informatics. His email is \email{jsheikh2@gmu.edu}.

\noindent {\bf CAT P. LE} received a B.S. degree in electrical and computer engineering from Rutgers University, an M.S. degree in electrical
engineering from the California Institute of Technology (Caltech), and a Ph.D. degree in electrical and computer engineering
from Duke University. His research interests include image processing, computer vision, and machine learning, with a focus
on transfer learning, and continual learning. His email address is \email{cat.le@duke.edu}. His website is \href{https://www.catphuocle.com/}{https://www.catphuocle.com/}

\noindent {\bf HODA BIDKHORI } is an assistant professor at the Department of Computational and Data Sciences at the George Mason University. She earned her Ph.D. in Applied Mathematics from the Massachusetts Institute of Technology (MIT), where she subsequently spent several years as a postdoctoral researcher and lecturer in Operations Research and Statistics. Her current research focuses on the theory and applications of data science. Her e-mail address is \email{hbidkhor@gmu.edu}. Her website is
\href{https://sites.google.com/view/hoda-bidkhori/home}{https://sites.google.com/view/hoda-bidkhori/home}.

\end{document}

%% file: wscbib.tex
%%%%%%%%%%%%%%%%%%%%%%%%%%%%%%%%%%%%%%%%%%%%%%%%%%%%%%%%%%%%%%%%%%%%%%%%%%%%%%
%                                                                            %
%     THESE COMMANDS ARE REQUIRED TO WORK WITH WSC.BST TO MAKE BIBLIO     %
%                                                                            %
%%%%%%%%%%%%%%%%%%%%%%%%%%%%%%%%%%%%%%%%%%%%%%%%%%%%%%%%%%%%%%%%%%%%%%%%%%%%%%
\makeatletter
\let\@internalcite\cite
\def\cite{\def\@citeseppen{-1000}%
    \def\@cite##1##2{(##1\if@tempswa , ##2\fi)}%
    \def\citeauthoryear##1##2##3{##1 ##3}\@internalcite}
\def\citeNP{\def\@citeseppen{-1000}%
    \def\@cite##1##2{##1\if@tempswa , ##2\fi}%
    \def\citeauthoryear##1##2##3{##1 ##3}\@internalcite}
\def\citeN{\def\@citeseppen{-1000}%
%  Pierre L'Ecuyer's fix for multiple cite bug
%  Added by Paul J Sanchez on 4 October 2001
%   \def\@cite##1##2{##1\if@tempswa , ##2)\else{)}\fi}%
%   \def\citeauthoryear##1##2##3{##1 (##3}\@citedata}
    \def\@cite##1##2{##1\if@tempswa, ##2)\else{}\fi}%
    \def\citeauthoryear##1##2##3{##1 (##3)}\@citedata}
\def\citeA{\def\@citeseppen{-1000}%
    \def\@cite##1##2{(##1\if@tempswa , ##2\fi)}%
    \def\citeauthoryear##1##2##3{##1}\@internalcite}
\def\citeANP{\def\@citeseppen{-1000}%
    \def\@cite##1##2{##1\if@tempswa , ##2\fi}%
    \def\citeauthoryear##1##2##3{##1}\@internalcite}
\def\shortcite{\def\@citeseppen{-1000}%
    \def\@cite##1##2{(##1\if@tempswa , ##2\fi)}%
    \def\citeauthoryear##1##2##3{##2 ##3}\@internalcite}
\def\shortciteNP{\def\@citeseppen{-1000}%
    \def\@cite##1##2{##1\if@tempswa , ##2\fi}%
    \def\citeauthoryear##1##2##3{##2 ##3}\@internalcite}
\def\shortciteN{\def\@citeseppen{-1000}%
%  Pierre L'Ecuyer's fix for multiple cite bug
%  Added by Paul J Sanchez on 2 September 2002
%  should have caught this last year...
%   \def\@cite##1##2{##1\if@tempswa , ##2)\else{)}\fi}%
%   \def\citeauthoryear##1##2##3{##2 (##3}\@citedata}
% Shane G. Henderson fix for extra right bracket at end of optional material June 8, 2005
%    \def\@cite##1##2{##1\if@tempswa, ##2)\else{}\fi}%
    \def\@cite##1##2{##1\if@tempswa, ##2\else{}\fi}%
    \def\citeauthoryear##1##2##3{##2 (##3)}\@citedata}
\def\shortciteA{\def\@citeseppen{-1000}%
    \def\@cite##1##2{(##1\if@tempswa , ##2\fi)}%
    \def\citeauthoryear##1##2##3{##2}\@internalcite}
\def\shortciteANP{\def\@citeseppen{-1000}%
    \def\@cite##1##2{##1\if@tempswa , ##2\fi}%
    \def\citeauthoryear##1##2##3{##2}\@internalcite}
\def\citeyear{\def\@citeseppen{-1000}%
    \def\@cite##1##2{(##1\if@tempswa , ##2\fi)}%
    \def\citeauthoryear##1##2##3{##3}\@citedata}
\def\citeyearNP{\def\@citeseppen{-1000}%
    \def\@cite##1##2{##1\if@tempswa , ##2\fi}%
    \def\citeauthoryear##1##2##3{##3}\@citedata}
%
% \@citedata and \@citedatax:
%
% Place commas in-between citations in the same \citeyear, \citeyearNP,
% \citeN, or \shortciteN command.
% Use something like \citeN{ref1,ref2,ref3} and \citeN{ref4} for a list.
%
\def\@citedata{%
    \@ifnextchar [{\@tempswatrue\@citedatax}%
                  {\@tempswafalse\@citedatax[]}%
}

\def\@citedatax[#1]#2{%
\if@filesw\immediate\write\@auxout{\string\citation{#2}}\fi%
  \def\@citea{}\@cite{\@for\@citeb:=#2\do%
    {\@citea\def\@citea{, }\@ifundefined% by Young
       {b@\@citeb}{{\bf ?}%
       \@warning{Citation `\@citeb' on page \thepage \space undefined}}%
{\csname b@\@citeb\endcsname}}}{#1}}%

% don't box citations, separate with ; and a space
% also, make the penalty between citations negative: a good place to break.
%
\def\@citex[#1]#2{%
\if@filesw\immediate\write\@auxout{\string\citation{#2}}\fi%
  \def\@citea{}\@cite{\@for\@citeb:=#2\do%
    {\@citea\def\@citea{; }\@ifundefined% by Young
       {b@\@citeb}{{\bf ?}%
       \@warning{Citation `\@citeb' on page \thepage \space undefined}}%
{\csname b@\@citeb\endcsname}}}{#1}}%

% (from apalike.sty)
% No labels in the bibliography.
%
\def\@biblabel#1{}
\makeatother

%\newlength{\bibhang}
%\setlength{\bibhang}{2em}

% Indent second and subsequent lines of bibliographic entries. Taken
% from openbib.sty: \newblock is set to {}.
% \renewcommand{\refname}{REFERENCES}

\newdimen\bibindent
\bibindent=0.0em
% SEC: was \def\thebibliography#1{\section*{\refname\@mkboth
% SEC: was   {\uppercase{\refname}}{\uppercase{\refname}}}\list
\def\thebibliography#1{\section*{\refname}\list
   {}{\settowidth\labelwidth{[#1]}
   \leftmargin\parindent
   \itemindent -\parindent
   \listparindent \itemindent
   \itemsep 0pt
   \parsep 0pt}
   \def\newblock{}
   \sloppy
   \sfcode`\.=1000\relax}